\documentclass[
]{ceurart}
\usepackage{multirow}
\usepackage[numbers]{natbib}
\usepackage{comment} 
\usepackage{hyperref}
\usepackage{graphics}
\usepackage{graphicx}
\usepackage{graphicx}
\usepackage{microtype}
\usepackage{caption, subcaption} 
\begin{document}

\copyrightyear{2020}
\copyrightclause{Copyright for this paper by its authors.
  Use permitted under Creative Commons License Attribution 4.0
  International (CC BY 4.0).}

\conference{FIRE 2021: Forum for Information Retrival Evaluation, December 13-17, 2021, India}

\title{IIITT@Dravidian-CodeMix-FIRE2021: Transliterate or translate? Sentiment analysis of code-mixed text in Dravidian languages}

\author[1]{Karthik Puranik}[%
orcid=0000-0002-0877-7063,
email=karthikp18c@iiitt.ac.in,
]
\address[1]{Indian Institute of Information Technology Tiruchirappalli}

\author[2]{Bharathi B}[%
orcid=0000-0001-7279-5357,
email=bharathib@ssn.edu.in,
]

\author[2]{Senthil Kumar B}[%
orcid=0000-0003-0835-5271,
email=senthil@ssn.edu.in,
]
\address[2]{Computer Science and Engineering, SSN College of Engineering, Chennai}

\begin{abstract}
Sentiment analysis of social media posts and comments for various marketing and emotional purposes is gaining recognition. With the increasing presence of code-mixed content in various native languages, there is a need for ardent research to produce promising results. This research paper bestows a tiny contribution to this research in the form of sentiment analysis of code-mixed social media comments in the popular Dravidian languages Kannada, Tamil and Malayalam. It describes the work for the shared task conducted by Dravidian-CodeMix at FIRE 2021 by employing pre-trained models like ULMFiT and multilingual BERT fine-tuned on the code-mixed dataset, transliteration (TRAI) of the same, English translations (TRAA) of the TRAI data and the combination of all the three. The results are recorded in this research paper where the best models stood 4th, 5th and 10th ranks in the Tamil, Kannada and Malayalam tasks respectively. 
\end{abstract}


\maketitle

\section{Introduction}
Sentiment analysis is a popular technique for analysing and evaluating textual content to learn the attitude and thoughts expressed in it \citep{RePEc:por:fepwps:489}. The term “Sentiment analysis” was first introduced in \citeauthor{inproceedings}. This method is largely employed in the marketing sector to realize the opinion of the customers on a certain product without reading all the feedbacks. Natural language processing (NLP) truly automates the wearisome tasks like analysing feedbacks. Several other tasks like sentiment classification, sentiment extraction, opinion summary, and subjectivity detection can also be performed \citep{keith2017hybrid} for various applications like spam email detection\citep{anees2020survey}, fake news detection \citep{9418446}, hate and hope speech detection \citep{puranik2021iiittltedieacl2021hope}, finding inappropriate texts in social media \citep{yasaswini-etal-2021-iiitt, jada-etal-2021-iiitt} and many others \citep{hande2021domain, hande2021benchmarking}. This paper concentrates on the sentiment analysis of code-mixed social media comments for Dravidian languages.

Social media is known to us as a virtual space to share our opinions, and communicate. However, social media is the largest hub for marketing. They provide spaces for brands to advertise products and target interested customers, which is the prime source of income for these platforms \citep{inbook}. In order to market the right product which appeals to its user, the social media platforms monitor their activities and comments\citep{chakravarthi-etal-2020-sentiment, chakravarthi-etal-2020-corpus}. This enables them to know the user's sentiment towards a product and company \citep{6425642}. Another crucial application of sentiment analysis is to automatically spot comments or posts which are offensive, abusive or spreads hatred in the social media platforms \citep{chakravarthi2020hopeedi}. Social media is a free space and no restrictions can be imposed on the comments or posts being circulated. However, the comments can certainly be detected and overseen to protect underage and the users who are vulnerable to get offended \citep{chen2012detecting, hegde2021images}. 

Social media features multilingual speakers from all over the world, and people tend to use a lot of variations while expressing their thoughts\citep{barman2014code}. Native speakers writing in Roman script is the most common scene due to the easy accessibility and customary usage of Roman script keyboards in mobile phones and desktop keyboards\cite{hande2021hope}. However, some users tend to write in the native script too. Finally, there is a case of code-mixing where two or more languages are merged in respect to the script or the usage \citep{8554413}. Sentiment analysis becomes difficult for such texts. In this paper, the method of transliterating the text is applied. Transliterating is the process of converting a text from one script to another while maintaining the pronunciation \citep{regmi2010understanding, Kalyan2021IIITTAC}. This brings about a uniformity in the text and helps the model learn better. However, due to the presence of English text in the code-mixed dataset, there has also been a slight effort to translate \cite{bahdanau2014neural} the text transliterated in the native language to English and train the model with it. 

This research paper depicts our work for the shared task Dravidian-CodeMix\footnote{\url{https://dravidian-codemix.github.io/2021/index.html}} at FIRE 2021 \cite{dravidiancodemix-findings, dravidiancodemix2021-overview}. The task was to detect the sentiment in the sentences for three of the major Dravidian languages \citep{krishnamurti2003dravidian} Kannada, Tamil, and Malayalam. Our system models stood 5th, 4th and 10th respectively in the shared task. The codes for the model and the transliterated and translated datasets are provided in the link\footnote{\url{https://github.com/karthikpuranik11/FIRE2021}}.

\section{Dataset}

The dataset provided by the organizers of the shared task has been used to train the models \citep{chakravarthi-etal-2020-sentiment, chakravarthi-etal-2020-corpus, hande-etal-2020-kancmd, chakravarthi-2020-hopeedi}. It contains annotated sentences obtained by cleaning YouTube comments\footnote{\url{https://www.youtube.com/}}. The sentences are highly code-mixed and contains inter-sentimental, intra-sentimental and tag switching which are prevalent in code-mixed data to be classified into five classes namely, positive, negative, unknown state, mixed feelings and not the intended language. The train, development and test distribution can be viewed in Table \ref{tab1}.

\begin{table}[!h]
\begin{center}
\renewcommand{\tabcolsep}{3mm}
 \scalebox{0.9}{
\begin{tabular}{|l|r|r|r|}
\hline

Split &  Kannada & Tamil & Malayalam\\
\hline
Training &  6,213 & 35,657 & 15,889\\
Development & 692 & 3,963 & 1,767\\
Test & 768 & 4,403 & 1,963\\
\hline
Total & 7,673 & 44,023 & 19,619\\
\hline
\end{tabular}
}
\end{center}
\caption{ Train-Development-Test Data Distribution}\label{tab1}
\end{table}

Further, the transliterations (TRAI) of the code-mixed training (TRA) dataset in the respective Dravidian languages were used. Small preprocessing steps like removing the language tag and brackets, removing all the sentences which belong to “not-language” were removed. The English translations (TRAA) of these transliterations was also used as a part of this research. It was evident that English was the most widely used language after the Dravidian language. A few comments represented in the TRA, TRAI and TRAA datasets belonging to the five classes are tabulated in Table \ref{fig1}.

\begin{table}[!h]
    \centering
    \includegraphics[width=1\linewidth, height=17cm]{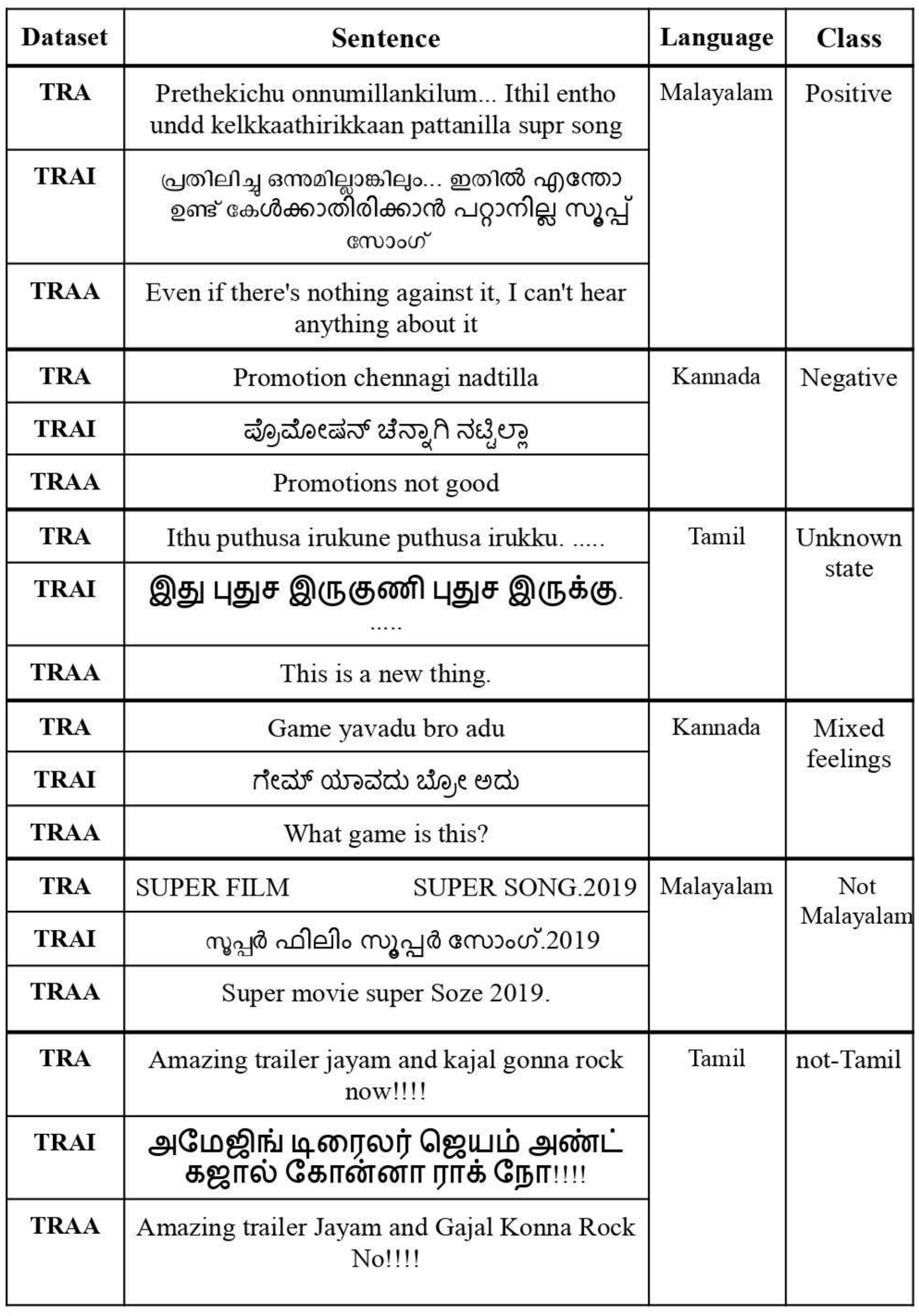}
    \caption{Examples of the code-mixed sentences, its transliteration and translation in Kannada, Tamil and Malayalam}
    \label{fig1}
\end{table}

\section{Methodology}

Based on previous researches, two of the most promising pre-trained models, ULMFiT \cite{howard2018universal} and BERT \cite{devlin2019bert} with bidirectional LSTM layers \cite{chiu2016named}, were used to determine the sentiment of the sentences. These models were fine-tuned separately on the training data provided by the organizers, the transliterated data combined with the training data, the translated data combined with the training data and the combination of all the three datasets.

\subsection{BERT}

Bidirectional Encoder Representation
from Transformers (BERT) is one of the most popular transformer based models, trained extensively on the entire Wikipedia and 0.11 million WordPiece sentences \cite{wu2016googles} for over 104 languages in the world. The unprecedented methods like Next Sentence Prediction (NSP) and Masked Language Modelling (MLM) successfully catch a deeper context of the languages. For the particular task, \textbf{\textit{bert-base-multilingual-cased}} \cite{pires-etal-2019-multilingual} from HuggingFace\footnote{\url{https://huggingface.co/}} \cite{wolf2020huggingfaces} has been used. It comprises 12 layers and attention heads and about 110M parameters. 

This model was further concatenated with bidirectional LSTM layers, which are known to improve the information being fed. The bidirectional layers read the embeddings from both the directions, hence, boosts the context and the F1 scores drastically. Further, the training was done with an Adam optimizer \cite{kingma2017adam}, a learning rate of 2e-5 with the \textit{cross-entropy} loss function \cite{zhang2018generalized, agarap2019deep} for a total of 5 epochs. The various parameters employed in the BERT+ BiLSTM model van be viewed in Table \ref{params}.

\begin{table}[htbp]
 
\begin{center}
    
\renewcommand{\tabcolsep}{3mm}
\scalebox{1}{
\begin{tabular}{|l r|} 
 \hline
 \textbf{Parameter}& \textbf{Value}\\ 
 \hline
 
Number of LSTM units & 256 \\ 
Dropout & 0.4 \\
Activation Function & ReLU \\
Max Len & 128 \\
Batch Size & 32 \\
Optimizer & AdamW\\
Learning Rate & 2e-5\\
Loss Function & cross-entropy\\
Number of epochs & 5\\
 
\hline
\end{tabular}
}
\end{center}
\caption{Parameters for the BERT+BiLSTM model. }
\label{params}
\end{table}

\subsection{ULMFiT}
Universal Language Model Fine-tuning, or ULMFiT was one of the initial transfer learning method to produce state-of-the-art results for NLP tasks. It was trained on very huge datasets like Wikitext-103\footnote{\url{https://blog.einstein.ai/the-wikitext-long-term-dependency-language-modeling-dataset/}} with around 103M sentences. It employs three novel techniques for fine-tuning the language models for various NLP tasks, which are discriminative fine-tuning, slanted triangular learning rates (STLR) and gradual unfreezing. AWD-LSTM language model \cite{merity2017regularizing, hande2021offensive}, a standard LSTM consisting 3 layers and 1150 hidden activation per
layer and an embedding size of 400 and without any attentions and just well tune dropouts, is generally used. Adam optimizer with starting learning rate of 1e-8 and an end learning rate of 1e-2 and a dropout of 0.5 is used.

\subsection{Transliteration}
The IndianNLP-Transliteration\footnote{\url{https://github.com/AI4Bharat/IndianNLP-Transliteration}} tool from AI4Bharat was used to get the transliterations of the training dataset. This deep transliteration tool can transliterate from Roman script to any low resourced Indian language. The architecture majorly consists of Recurrent Neural Networks (RNN) \citep{Sherstinsky_2020} with encoders and decoders \cite{vaswani2017attention}. The decoder employs top 'k' predictions and then re-ranked to get the most probable word \citep{bahdanau2016neural}. It is observed that most of the sentences in the Dravidian language present in the code-mixed dataset is the languages written in Roman script. The multilingual pre-trained models, largely trained on these Dravidian languages in their original scripts, might find it hard to comprehend such sentences. Transliterating them back to the original script could possibly improve the accuracy \citep{thomas2020sentimental}.

\subsection{Translation}
The transliterated data in the Dravidian language is translated to English using IndicTrans \cite{ramesh2021samanantar} from AI4Bharat\footnote{\url{https://github.com/AI4Bharat/indicTrans}}. This PyTorch Fairseq\footnote{\url{https://github.com/pytorch/fairseq}} \citep{ott2019fairseq, Puranik2021AttentiveFO} based Transformer NMT model, is trained on a large parallel corpus containing 46.9 million sentences of Samanantar dataset. The model is known to produce state-of-the-art BLEU \citep{article23} scores for 11 Indian languages. The translations given by the IndicTrans baseline model on the transliterated dataset was used. The reason for using the translated data was due to the presence of excessive English in the code-mixed dataset, and most of the pre-trained models are trained on large number of English sentences.

\section{Results}

In this section, the F1 scores of the BERT and ULMFiT models for the sentiment analysis of Kannada, Tamil and Malayalam datasets are compared, and suitable analysis are recorded. The weighted F1 scores are tabulated in Table \ref{tbl2}. The models are fine-tuned on training dataset (TRA), the combination of transliterated dataset and TRA (TRAI), translated (TRAA) dataset and TRA and all 3 merged.

\begin{table}[width=.9\linewidth,cols=4,pos=!h]
\caption{Weighted F1-scores of sentiment analysis on the test datasets, where P: Precision, R: Recall and F1: F1 score.}\label{tbl2}
\begin{tabular*}{\tblwidth}{@{} L|RRR|RRR@{} }
\toprule
\textbf{Dataset} &\multicolumn{6}{c}{\textbf{Kannada}} \\
\midrule
& \multicolumn{3}{c|}{BERT} & \multicolumn{3}{c}{ULMFiT} \\ \midrule
&\textbf{P} & \textbf{R} & \textbf{F1} & \textbf{P} & \textbf{R} & \textbf{F1}  \\  \midrule
Train (TRA)&0.5952&0.6185&0.6040&\textbf{0.6547}&\textbf{0.6276} & \textbf{0.6389}\\
Transliterate + TRA (TRAI)&0.5587  &  0.6133   & 0.5831&  0.6239  &  0.6081   & 0.6150 \\ 
Translate + TRA (TRAA)&0.6176   & 0.6367  &  0.6231& 0.6078 &   0.5990  &  0.6031\\
Merged (TRA+TRAI+TRAA)&0.6079 &   0.6172   & 0.6113& 0.6172   & 0.5885   & 0.5993\\
\midrule
& \multicolumn{6}{c}{\textbf{Tamil}} \\ \midrule
& \multicolumn{3}{c|}{BERT}  & \multicolumn{3}{c}{ULMFiT}\\ \midrule
&\textbf{P} & \textbf{R} & \textbf{F1} & \textbf{P} & \textbf{R} & \textbf{F1}  \\  \midrule
Train (TRA)&0.5291  &  0.5572   & 0.5308& 0.6544  &  0.6229  &  0.6362\\
Transliterate + TRA (TRAI)& 0.5334   & 0.5502  &  0.5366& \textbf{0.6889}  &  \textbf{0.6372}   & \textbf{0.6583}\\ 
Translate + TRA (TRAA)& 0.5284   & 0.5427   & 0.5310& 0.6694   & 0.6379  &  0.6514\\
Merged (TRA+TRAI+TRAA)&0.5298  &  0.5570   & 0.5367& 0.6629&    0.6306   & 0.6432\\
\midrule
& \multicolumn{6}{c}{\textbf{Malayalam}}\\ \midrule
& \multicolumn{3}{c|}{BERT} & \multicolumn{3}{c}{ULMFiT} \\ \midrule
&\textbf{P} & \textbf{R} & \textbf{F1} & \textbf{P} & \textbf{R} & \textbf{F1} \\  \midrule
Train (TRA)& 0.6238   & 0.6733  &  0.6457& 0.7084 &   0.6937   & 0.6990\\
Transliterate + TRA (TRAI)& 0.6874   & 0.7018  &  0.6933& \textbf{0.7139}   & \textbf{0.7013}  &  \textbf{0.7062}\\ 
Translate + TRA (TRAA)& 0.5976   & 0.7142  &  0.6467 & 0.7086   & 0.6901   & 0.6970\\
Merged (TRA+TRAI+TRAA) & 0.6822  &  0.6927   & 0.6863& 0.7041&    0.6952  &  0.6984\\
\bottomrule
\end{tabular*}
\end{table}

It is firstly clear from Table \ref{tbl2} that ULMFiT manages to get better F1 scores than BERT concatenated with biLSTM layers for the majority of the datasets. The unique transfer learning techniques used by ULMFiT like the discriminative fine-tuning, slanted triangular learning rates and gradual unfreezing seem to successfully produce exceptional F1 scores. Discriminative fine-tuning allows us to fine-tune each layer separately with different learning rates. Gradual unfreezing improves it further by keeping the last layer frozen in the first epoch and unfreezing layer by layer for the further epochs. Except for the Tamil data, BERT manages to give results comparable to ULMFiT for the other languages. 

ULMFiT fine-tuned on the TRA dataset gives the best F1-score of 0.639 for the Kannada task. It is followed by BERT fine-tuned on the TRAA dataset with 0.623. Other models gave similar results. It is surprising how the models managed to give F1 scores akin to other languages, considering the limited size of the dataset. ULMFiT manages to surpass BERT by a huge difference for the Tamil task. The presence of class imbalances in the Tamil dataset could be a reason for this issue. The “positive” comments are 2,830 in number out of the 4,402 sentences in the test dataset, while “not-Tamil” which is just 210. This imbalance causes a variation in the results. ULMFiT on TRAI and TRAA gave nearly similar F1 scores of 0.658 and 0.651 respectively. ULMFiT trained on all the four datasets gave equivalent results for the Malayalam task, with TRAI giving the best score of 0.706. BERT trained on TRAI gave a competitive score of 0.6933 for the same task. 

The basic observation derived while comparing the various datasets is the equal contention between the four datasets used. But, the most common scenario is that the TRAI dataset manages to have the upper hand in the majority of the cases. The most plausible explanation to this is due to the fact that the dataset is code-mixed and the Dravidian text written in Roman script. When that is converted to the native script, the model manages to fine-tune well. With the original data also present, the model manages to fine-tune on the English text too. However, we can't be entirely sure of the accuracy of transliterations from the IndianNLP-Transliteration tool. Another drawback of transliterating the code-mixed sentences is that the English and other language also get transliterated to the Kannada/Tamil/Malayalam script. Such words might not be able to recognized by the model at all. Dravidian languages can be complex and there might be several ways in which the comments in the Roman script can be transliterated, and a slight variation can change the meaning entirely\cite{inproceedings25}. However, to tackle these, we merge the transliterated dataset with the TRA data so that the model manages to learn the other languages in the code-mixed data too. 

The TRAA and the merged dataset proves to be inefficient due to its low F1 scores. The TRAA dataset is not significantly behind the TRAI data, which proves that there is a scope to increase the accuracy with further research. Though IndicTrans one of the best models for machine translation of Indian languages has been employed, we can surely not rely entirely on the translations of the transliterated data. Further, cleaning of the TRAA data by removing sentences which fail to make any sense and fine-tuning the IndicTrans model on a suitable parallel corpus before translating it can be done to obtain good F1 scores for the TRAA dataset. The combination of the three datasets however fails miserably in most of the cases due to the repetition of the sentences in different forms, which seems to make the model not learning anything productively, and the inaccuracies in the TRAA and TRAI datasets add up to reduce the F1 scores even further.

\section{Conclusion}
Sentiment analysis of social media comments emerges as one of the most notable tasks of natural language processing (NLP). In order to obtain good F1 scores for the sentiment analysis of social media comments in code-mixed Dravidian languages Kannada, Tamil and Malayalam, after careful experimentation with Transformer based ULMFiT and mBERT fine-tuned on TRA, TRAI, TRAA and merged dataset, ULMFiT proved to give the best F1 scores for all the three languages. For Kannada, it was on the TRA dataset, while TRAI proved effective for Tamil and Malayalam. This paper introduces the use of TRAA dataset which can be worked upon in the future.

\bibliography{ceu}

\begin{thebibliography}{48}
\expandafter\ifx\csname natexlab\endcsname\relax\def\natexlab#1{#1}\fi
\providecommand{\url}[1]{\texttt{#1}}
\providecommand{\href}[2]{#2}
\providecommand{\path}[1]{#1}
\providecommand{\DOIprefix}{doi:}
\providecommand{\ArXivprefix}{arXiv:}
\providecommand{\URLprefix}{URL: }
\providecommand{\Pubmedprefix}{pmid:}
\providecommand{\doi}[1]{\href{http://dx.doi.org/#1}{\path{#1}}}
\providecommand{\Pubmed}[1]{\href{pmid:#1}{\path{#1}}}
\providecommand{\bibinfo}[2]{#2}
\ifx\xfnm\relax \def\xfnm[#1]{\unskip,\space#1}\fi
\bibitem[{Rambocas and Gama(2013)}]{RePEc:por:fepwps:489}
\bibinfo{author}{M.~Rambocas}, \bibinfo{author}{J.~Gama},
  \bibinfo{title}{{Marketing Research: The Role Of Sentiment Analysis}},
  \bibinfo{type}{FEP Working Papers} \bibinfo{number}{489}, Universidade do
  Porto, Faculdade de Economia do Porto, \bibinfo{year}{2013}. \URLprefix
  \url{https://ideas.repec.org/p/por/fepwps/489.html}.
\bibitem[{Nasukawa and Yi(2003)}]{inproceedings}
\bibinfo{author}{T.~Nasukawa}, \bibinfo{author}{J.~Yi},
\newblock \bibinfo{title}{Sentiment analysis: Capturing favorability using
  natural language processing},
\newblock \bibinfo{year}{2003}, pp. \bibinfo{pages}{70--77}.
  \DOIprefix\doi{10.1145/945645.945658}.
\bibitem[{Keith et~al.(2017)Keith, Fuentes, and Meneses}]{keith2017hybrid}
\bibinfo{author}{B.~Keith}, \bibinfo{author}{E.~Fuentes},
  \bibinfo{author}{C.~Meneses},
\newblock \bibinfo{title}{A hybrid approach for sentiment analysis applied to
  paper},
\newblock in: \bibinfo{booktitle}{Proceedings of ACM SIGKDD Conference,
  Halifax, Nova Scotia, Canada}, \bibinfo{year}{2017}, p.~\bibinfo{pages}{10}.
\bibitem[{Anees et~al.(2020)Anees, Shaikh, Shaikh, and
  Shaikh}]{anees2020survey}
\bibinfo{author}{A.~F. Anees}, \bibinfo{author}{A.~Shaikh},
  \bibinfo{author}{A.~Shaikh}, \bibinfo{author}{S.~Shaikh},
\newblock \bibinfo{title}{Survey paper on sentiment analysis: Techniques and
  challenges},
\newblock \bibinfo{journal}{EasyChair2516-2314}  (\bibinfo{year}{2020}).
\bibitem[{Hande et~al.(2021)Hande, Puranik, Priyadharshini, Thavareesan, and
  Chakravarthi}]{9418446}
\bibinfo{author}{A.~Hande}, \bibinfo{author}{K.~Puranik},
  \bibinfo{author}{R.~Priyadharshini}, \bibinfo{author}{S.~Thavareesan},
  \bibinfo{author}{B.~R. Chakravarthi},
\newblock \bibinfo{title}{Evaluating pretrained transformer-based models for
  covid-19 fake news detection},
\newblock in: \bibinfo{booktitle}{2021 5th International Conference on
  Computing Methodologies and Communication (ICCMC)}, \bibinfo{year}{2021}, pp.
  \bibinfo{pages}{766--772}. \DOIprefix\doi{10.1109/ICCMC51019.2021.9418446}.
\bibitem[{Puranik et~al.(2021)Puranik, Hande, Priyadharshini, Thavareesan, and
  Chakravarthi}]{puranik2021iiittltedieacl2021hope}
\bibinfo{author}{K.~Puranik}, \bibinfo{author}{A.~Hande},
  \bibinfo{author}{R.~Priyadharshini}, \bibinfo{author}{S.~Thavareesan},
  \bibinfo{author}{B.~R. Chakravarthi},
  \bibinfo{title}{Iiitt@lt-edi-eacl2021-hope speech detection: There is always
  hope in transformers}, \bibinfo{year}{2021}.
  \href{http://arxiv.org/abs/2104.09066}{{\tt arXiv:2104.09066}}.
\bibitem[{Yasaswini et~al.(2021)Yasaswini, Puranik, Hande, Priyadharshini,
  Thavareesan, and Chakravarthi}]{yasaswini-etal-2021-iiitt}
\bibinfo{author}{K.~Yasaswini}, \bibinfo{author}{K.~Puranik},
  \bibinfo{author}{A.~Hande}, \bibinfo{author}{R.~Priyadharshini},
  \bibinfo{author}{S.~Thavareesan}, \bibinfo{author}{B.~R. Chakravarthi},
\newblock \bibinfo{title}{{IIITT}@{D}ravidian{L}ang{T}ech-{EACL}2021: Transfer
  learning for offensive language detection in {D}ravidian languages},
\newblock in: \bibinfo{booktitle}{Proceedings of the First Workshop on Speech
  and Language Technologies for Dravidian Languages},
  \bibinfo{publisher}{Association for Computational Linguistics},
  \bibinfo{address}{Kyiv}, \bibinfo{year}{2021}, pp. \bibinfo{pages}{187--194}.
  \URLprefix \url{https://aclanthology.org/2021.dravidianlangtech-1.25}.
\bibitem[{Jada et~al.(2021)Jada, Reddy, Yasaswini, Prabakaran, Sampath, and
  Thangasamy}]{jada-etal-2021-iiitt}
\bibinfo{author}{P.~K. Jada}, \bibinfo{author}{D.~S. Reddy},
  \bibinfo{author}{K.~Yasaswini}, \bibinfo{author}{C.~Prabakaran},
  \bibinfo{author}{A.~Sampath}, \bibinfo{author}{S.~Thangasamy},
\newblock \bibinfo{title}{{IIIT@Dravidian-CodeMix-FIRE2021: Transformer Model
  based Sentiment Analysis in Dravidian Languages}},
\newblock in: \bibinfo{booktitle}{Working Notes of FIRE 2021 - Forum for
  Information Retrieval Evaluation}, \bibinfo{publisher}{CEUR},
  \bibinfo{year}{2021}.
\bibitem[{Hande et~al.(2021{\natexlab{a}})Hande, Puranik, Priyadharshini, and
  Chakravarthi}]{hande2021domain}
\bibinfo{author}{A.~Hande}, \bibinfo{author}{K.~Puranik},
  \bibinfo{author}{R.~Priyadharshini}, \bibinfo{author}{B.~R. Chakravarthi},
\newblock \bibinfo{title}{Domain identification of scientific articles using
  transfer learning and ensembles},
\newblock in: \bibinfo{booktitle}{Trends and Applications in Knowledge
  Discovery and Data Mining: PAKDD 2021 Workshops, WSPA, MLMEIN, SDPRA, DARAI,
  and AI4EPT, Delhi, India, May 11, 2021 Proceedings 25},
  \bibinfo{organization}{Springer International Publishing},
  \bibinfo{year}{2021}{\natexlab{a}}, pp. \bibinfo{pages}{88--97}.
\bibitem[{Hande et~al.(2021{\natexlab{b}})Hande, Hegde, Priyadharshini,
  Ponnusamy, Kumaresan, Thavareesan, and Chakravarthi}]{hande2021benchmarking}
\bibinfo{author}{A.~Hande}, \bibinfo{author}{S.~U. Hegde},
  \bibinfo{author}{R.~Priyadharshini}, \bibinfo{author}{R.~Ponnusamy},
  \bibinfo{author}{P.~K. Kumaresan}, \bibinfo{author}{S.~Thavareesan},
  \bibinfo{author}{B.~R. Chakravarthi}, \bibinfo{title}{Benchmarking multi-task
  learning for sentiment analysis and offensive language identification in
  under-resourced dravidian languages}, \bibinfo{year}{2021}{\natexlab{b}}.
  \href{http://arxiv.org/abs/2108.03867}{{\tt arXiv:2108.03867}}.
\bibitem[{Oikonomidis and Fouskas(2019)}]{inbook}
\bibinfo{author}{T.~Oikonomidis}, \bibinfo{author}{K.~Fouskas},
  \bibinfo{title}{Is Social Media Paying Its Money?}, \bibinfo{year}{2019}, pp.
  \bibinfo{pages}{999--1006}. \DOIprefix\doi{10.1007/978-3-030-12453-3_115}.
\bibitem[{Chakravarthi et~al.(2020{\natexlab{a}})Chakravarthi, Jose,
  Suryawanshi, Sherly, and McCrae}]{chakravarthi-etal-2020-sentiment}
\bibinfo{author}{B.~R. Chakravarthi}, \bibinfo{author}{N.~Jose},
  \bibinfo{author}{S.~Suryawanshi}, \bibinfo{author}{E.~Sherly},
  \bibinfo{author}{J.~P. McCrae},
\newblock \bibinfo{title}{A sentiment analysis dataset for code-mixed
  {M}alayalam-{E}nglish},
\newblock in: \bibinfo{booktitle}{Proceedings of the 1st Joint Workshop on
  Spoken Language Technologies for Under-resourced languages (SLTU) and
  Collaboration and Computing for Under-Resourced Languages (CCURL)},
  \bibinfo{publisher}{European Language Resources association},
  \bibinfo{address}{Marseille, France}, \bibinfo{year}{2020}{\natexlab{a}}, pp.
  \bibinfo{pages}{177--184}. \URLprefix
  \url{https://www.aclweb.org/anthology/2020.sltu-1.25}.
\bibitem[{Chakravarthi et~al.(2020{\natexlab{b}})Chakravarthi, Muralidaran,
  Priyadharshini, and McCrae}]{chakravarthi-etal-2020-corpus}
\bibinfo{author}{B.~R. Chakravarthi}, \bibinfo{author}{V.~Muralidaran},
  \bibinfo{author}{R.~Priyadharshini}, \bibinfo{author}{J.~P. McCrae},
\newblock \bibinfo{title}{Corpus creation for sentiment analysis in code-mixed
  {T}amil-{E}nglish text},
\newblock in: \bibinfo{booktitle}{Proceedings of the 1st Joint Workshop on
  Spoken Language Technologies for Under-resourced languages (SLTU) and
  Collaboration and Computing for Under-Resourced Languages (CCURL)},
  \bibinfo{publisher}{European Language Resources association},
  \bibinfo{address}{Marseille, France}, \bibinfo{year}{2020}{\natexlab{b}}, pp.
  \bibinfo{pages}{202--210}. \URLprefix
  \url{https://www.aclweb.org/anthology/2020.sltu-1.28}.
\bibitem[{Neri et~al.(2012)Neri, Aliprandi, Capeci, Cuadros, and By}]{6425642}
\bibinfo{author}{F.~Neri}, \bibinfo{author}{C.~Aliprandi},
  \bibinfo{author}{F.~Capeci}, \bibinfo{author}{M.~Cuadros},
  \bibinfo{author}{T.~By},
\newblock \bibinfo{title}{Sentiment analysis on social media},
\newblock in: \bibinfo{booktitle}{2012 IEEE/ACM International Conference on
  Advances in Social Networks Analysis and Mining}, \bibinfo{year}{2012}, pp.
  \bibinfo{pages}{919--926}. \DOIprefix\doi{10.1109/ASONAM.2012.164}.
\bibitem[{Chakravarthi(2020)}]{chakravarthi2020hopeedi}
\bibinfo{author}{B.~R. Chakravarthi},
\newblock \bibinfo{title}{Hopeedi: A multilingual hope speech detection dataset
  for equality, diversity, and inclusion},
\newblock in: \bibinfo{booktitle}{Proceedings of the Third Workshop on
  Computational Modeling of People's Opinions, Personality, and Emotion's in
  Social Media}, \bibinfo{year}{2020}, pp. \bibinfo{pages}{41--53}.
\bibitem[{Chen et~al.(2012)Chen, Zhou, Zhu, and Xu}]{chen2012detecting}
\bibinfo{author}{Y.~Chen}, \bibinfo{author}{Y.~Zhou}, \bibinfo{author}{S.~Zhu},
  \bibinfo{author}{H.~Xu},
\newblock \bibinfo{title}{Detecting offensive language in social media to
  protect adolescent online safety},
\newblock in: \bibinfo{booktitle}{2012 International Conference on Privacy,
  Security, Risk and Trust and 2012 International Confernece on Social
  Computing}, \bibinfo{organization}{IEEE}, \bibinfo{year}{2012}, pp.
  \bibinfo{pages}{71--80}.
\bibitem[{Hegde et~al.(2021)Hegde, Hande, Priyadharshini, Thavareesan,
  Sakuntharaj, Thangasamy, Bharathi, and Chakravarthi}]{hegde2021images}
\bibinfo{author}{S.~U. Hegde}, \bibinfo{author}{A.~Hande},
  \bibinfo{author}{R.~Priyadharshini}, \bibinfo{author}{S.~Thavareesan},
  \bibinfo{author}{R.~Sakuntharaj}, \bibinfo{author}{S.~Thangasamy},
  \bibinfo{author}{B.~Bharathi}, \bibinfo{author}{B.~R. Chakravarthi},
  \bibinfo{title}{Do images really do the talking? analysing the significance
  of images in tamil troll meme classification}, \bibinfo{year}{2021}.
  \href{http://arxiv.org/abs/2108.03886}{{\tt arXiv:2108.03886}}.
\bibitem[{Barman et~al.(2014)Barman, Das, Wagner, and Foster}]{barman2014code}
\bibinfo{author}{U.~Barman}, \bibinfo{author}{A.~Das},
  \bibinfo{author}{J.~Wagner}, \bibinfo{author}{J.~Foster},
\newblock \bibinfo{title}{Code mixing: A challenge for language identification
  in the language of social media},
\newblock in: \bibinfo{booktitle}{Proceedings of the first workshop on
  computational approaches to code switching}, \bibinfo{year}{2014}, pp.
  \bibinfo{pages}{13--23}.
\bibitem[{Hande et~al.(2021)Hande, Priyadharshini, Sampath, Thamburaj,
  Chandran, and Chakravarthi}]{hande2021hope}
\bibinfo{author}{A.~Hande}, \bibinfo{author}{R.~Priyadharshini},
  \bibinfo{author}{A.~Sampath}, \bibinfo{author}{K.~P. Thamburaj},
  \bibinfo{author}{P.~Chandran}, \bibinfo{author}{B.~R. Chakravarthi},
  \bibinfo{title}{Hope speech detection in under-resourced kannada language},
  \bibinfo{year}{2021}. \href{http://arxiv.org/abs/2108.04616}{{\tt
  arXiv:2108.04616}}.
\bibitem[{Thara and Poornachandran(2018)}]{8554413}
\bibinfo{author}{S.~Thara}, \bibinfo{author}{P.~Poornachandran},
\newblock \bibinfo{title}{Code-mixing: A brief survey},
\newblock in: \bibinfo{booktitle}{2018 International Conference on Advances in
  Computing, Communications and Informatics (ICACCI)}, \bibinfo{year}{2018},
  pp. \bibinfo{pages}{2382--2388}. \DOIprefix\doi{10.1109/ICACCI.2018.8554413}.
\bibitem[{Regmi et~al.(2010)Regmi, Naidoo, and
  Pilkington}]{regmi2010understanding}
\bibinfo{author}{K.~Regmi}, \bibinfo{author}{J.~Naidoo},
  \bibinfo{author}{P.~Pilkington},
\newblock \bibinfo{title}{Understanding the processes of translation and
  transliteration in qualitative research},
\newblock \bibinfo{journal}{International Journal of Qualitative Methods}
  \bibinfo{volume}{9} (\bibinfo{year}{2010}) \bibinfo{pages}{16--26}.
\bibitem[{Kalyan et~al.(2021)Kalyan, Reddy, Hande, Priyadharshini, Sakuntharaj,
  and Chakravarthi}]{Kalyan2021IIITTAC}
\bibinfo{author}{P.~Kalyan}, \bibinfo{author}{D.~Reddy},
  \bibinfo{author}{A.~Hande}, \bibinfo{author}{R.~Priyadharshini},
  \bibinfo{author}{R.~Sakuntharaj}, \bibinfo{author}{B.~R. Chakravarthi},
\newblock \bibinfo{title}{Iiitt at case 2021 task 1: Leveraging pretrained
  language models for multilingual protest detection},
\newblock in: \bibinfo{booktitle}{CASE}, \bibinfo{year}{2021}.
\bibitem[{Bahdanau et~al.(2014)Bahdanau, Cho, and Bengio}]{bahdanau2014neural}
\bibinfo{author}{D.~Bahdanau}, \bibinfo{author}{K.~Cho},
  \bibinfo{author}{Y.~Bengio},
\newblock \bibinfo{title}{Neural machine translation by jointly learning to
  align and translate},
\newblock \bibinfo{journal}{arXiv preprint arXiv:1409.0473}
  (\bibinfo{year}{2014}).
\bibitem[{Chakravarthi et~al.(2021)Chakravarthi, Priyadharshini, Thavareesan,
  Chinnappa, Thenmozhi, Sherly, McCrae, Hande, Ponnusamy, Banerjee, and
  Vasantharajan}]{dravidiancodemix-findings}
\bibinfo{author}{B.~R. Chakravarthi}, \bibinfo{author}{R.~Priyadharshini},
  \bibinfo{author}{S.~Thavareesan}, \bibinfo{author}{D.~Chinnappa},
  \bibinfo{author}{D.~Thenmozhi}, \bibinfo{author}{E.~Sherly},
  \bibinfo{author}{J.~P. McCrae}, \bibinfo{author}{A.~Hande},
  \bibinfo{author}{R.~Ponnusamy}, \bibinfo{author}{S.~Banerjee},
  \bibinfo{author}{C.~Vasantharajan},
\newblock \bibinfo{title}{{Findings of the Sentiment Analysis of Dravidian
  Languages in Code-Mixed Text}},
\newblock in: \bibinfo{booktitle}{Working Notes of FIRE 2021 - Forum for
  Information Retrieval Evaluation}, \bibinfo{publisher}{CEUR},
  \bibinfo{year}{2021}.
\bibitem[{Priyadharshini et~al.(2021)Priyadharshini, Chakravarthi, Thavareesan,
  Chinnappa, Thenmozi, and Sherly}]{dravidiancodemix2021-overview}
\bibinfo{author}{R.~Priyadharshini}, \bibinfo{author}{B.~R. Chakravarthi},
  \bibinfo{author}{S.~Thavareesan}, \bibinfo{author}{D.~Chinnappa},
  \bibinfo{author}{D.~Thenmozi}, \bibinfo{author}{E.~Sherly},
\newblock \bibinfo{title}{Overview of the dravidiancodemix 2021 shared task on
  sentiment detection in tamil, malayalam, and kannada},
\newblock in: \bibinfo{booktitle}{Forum for Information Retrieval Evaluation},
  FIRE 2021, \bibinfo{publisher}{Association for Computing Machinery},
  \bibinfo{year}{2021}.
\bibitem[{Krishnamurti(2003)}]{krishnamurti2003dravidian}
\bibinfo{author}{B.~Krishnamurti}, \bibinfo{title}{The dravidian languages},
  \bibinfo{publisher}{Cambridge University Press}, \bibinfo{year}{2003}.
\bibitem[{Hande et~al.(2020)Hande, Priyadharshini, and
  Chakravarthi}]{hande-etal-2020-kancmd}
\bibinfo{author}{A.~Hande}, \bibinfo{author}{R.~Priyadharshini},
  \bibinfo{author}{B.~R. Chakravarthi},
\newblock \bibinfo{title}{{K}an{CMD}: {K}annada {C}ode{M}ixed dataset for
  sentiment analysis and offensive language detection},
\newblock in: \bibinfo{booktitle}{Proceedings of the Third Workshop on
  Computational Modeling of People's Opinions, Personality, and Emotion's in
  Social Media}, \bibinfo{publisher}{Association for Computational
  Linguistics}, \bibinfo{address}{Barcelona, Spain (Online)},
  \bibinfo{year}{2020}, pp. \bibinfo{pages}{54--63}. \URLprefix
  \url{https://www.aclweb.org/anthology/2020.peoples-1.6}.
\bibitem[{Chakravarthi(2020)}]{chakravarthi-2020-hopeedi}
\bibinfo{author}{B.~R. Chakravarthi},
\newblock \bibinfo{title}{{H}ope{EDI}: A multilingual hope speech detection
  dataset for equality, diversity, and inclusion},
\newblock in: \bibinfo{booktitle}{Proceedings of the Third Workshop on
  Computational Modeling of People's Opinions, Personality, and Emotion's in
  Social Media}, \bibinfo{publisher}{Association for Computational
  Linguistics}, \bibinfo{address}{Barcelona, Spain (Online)},
  \bibinfo{year}{2020}, pp. \bibinfo{pages}{41--53}. \URLprefix
  \url{https://aclanthology.org/2020.peoples-1.5}.
\bibitem[{Howard and Ruder(2018)}]{howard2018universal}
\bibinfo{author}{J.~Howard}, \bibinfo{author}{S.~Ruder},
  \bibinfo{title}{Universal language model fine-tuning for text
  classification}, \bibinfo{year}{2018}.
  \href{http://arxiv.org/abs/1801.06146}{{\tt arXiv:1801.06146}}.
\bibitem[{Devlin et~al.(2019)Devlin, Chang, Lee, and
  Toutanova}]{devlin2019bert}
\bibinfo{author}{J.~Devlin}, \bibinfo{author}{M.-W. Chang},
  \bibinfo{author}{K.~Lee}, \bibinfo{author}{K.~Toutanova},
  \bibinfo{title}{Bert: Pre-training of deep bidirectional transformers for
  language understanding}, \bibinfo{year}{2019}.
  \href{http://arxiv.org/abs/1810.04805}{{\tt arXiv:1810.04805}}.
\bibitem[{Chiu and Nichols(2016)}]{chiu2016named}
\bibinfo{author}{J.~P.~C. Chiu}, \bibinfo{author}{E.~Nichols},
  \bibinfo{title}{Named entity recognition with bidirectional lstm-cnns},
  \bibinfo{year}{2016}. \href{http://arxiv.org/abs/1511.08308}{{\tt
  arXiv:1511.08308}}.
\bibitem[{Wu et~al.(2016)Wu, Schuster, Chen, Le, Norouzi, Macherey, Krikun,
  Cao, Gao, Macherey, Klingner, Shah, Johnson, Liu, Łukasz Kaiser, Gouws,
  Kato, Kudo, Kazawa, Stevens, Kurian, Patil, Wang, Young, Smith, Riesa,
  Rudnick, Vinyals, Corrado, Hughes, and Dean}]{wu2016googles}
\bibinfo{author}{Y.~Wu}, \bibinfo{author}{M.~Schuster},
  \bibinfo{author}{Z.~Chen}, \bibinfo{author}{Q.~V. Le},
  \bibinfo{author}{M.~Norouzi}, \bibinfo{author}{W.~Macherey},
  \bibinfo{author}{M.~Krikun}, \bibinfo{author}{Y.~Cao},
  \bibinfo{author}{Q.~Gao}, \bibinfo{author}{K.~Macherey},
  \bibinfo{author}{J.~Klingner}, \bibinfo{author}{A.~Shah},
  \bibinfo{author}{M.~Johnson}, \bibinfo{author}{X.~Liu},
  \bibinfo{author}{Łukasz Kaiser}, \bibinfo{author}{S.~Gouws},
  \bibinfo{author}{Y.~Kato}, \bibinfo{author}{T.~Kudo},
  \bibinfo{author}{H.~Kazawa}, \bibinfo{author}{K.~Stevens},
  \bibinfo{author}{G.~Kurian}, \bibinfo{author}{N.~Patil},
  \bibinfo{author}{W.~Wang}, \bibinfo{author}{C.~Young},
  \bibinfo{author}{J.~Smith}, \bibinfo{author}{J.~Riesa},
  \bibinfo{author}{A.~Rudnick}, \bibinfo{author}{O.~Vinyals},
  \bibinfo{author}{G.~Corrado}, \bibinfo{author}{M.~Hughes},
  \bibinfo{author}{J.~Dean}, \bibinfo{title}{Google's neural machine
  translation system: Bridging the gap between human and machine translation},
  \bibinfo{year}{2016}. \href{http://arxiv.org/abs/1609.08144}{{\tt
  arXiv:1609.08144}}.
\bibitem[{Pires et~al.(2019)Pires, Schlinger, and
  Garrette}]{pires-etal-2019-multilingual}
\bibinfo{author}{T.~Pires}, \bibinfo{author}{E.~Schlinger},
  \bibinfo{author}{D.~Garrette},
\newblock \bibinfo{title}{How multilingual is multilingual {BERT}?},
\newblock in: \bibinfo{booktitle}{Proceedings of the 57th Annual Meeting of the
  Association for Computational Linguistics}, \bibinfo{publisher}{Association
  for Computational Linguistics}, \bibinfo{address}{Florence, Italy},
  \bibinfo{year}{2019}, pp. \bibinfo{pages}{4996--5001}. \URLprefix
  \url{https://aclanthology.org/P19-1493}.
  \DOIprefix\doi{10.18653/v1/P19-1493}.
\bibitem[{Wolf et~al.(2020)Wolf, Debut, Sanh, Chaumond, Delangue, Moi, Cistac,
  Rault, Louf, Funtowicz, Davison, Shleifer, von Platen, Ma, Jernite, Plu, Xu,
  Scao, Gugger, Drame, Lhoest, and Rush}]{wolf2020huggingfaces}
\bibinfo{author}{T.~Wolf}, \bibinfo{author}{L.~Debut},
  \bibinfo{author}{V.~Sanh}, \bibinfo{author}{J.~Chaumond},
  \bibinfo{author}{C.~Delangue}, \bibinfo{author}{A.~Moi},
  \bibinfo{author}{P.~Cistac}, \bibinfo{author}{T.~Rault},
  \bibinfo{author}{R.~Louf}, \bibinfo{author}{M.~Funtowicz},
  \bibinfo{author}{J.~Davison}, \bibinfo{author}{S.~Shleifer},
  \bibinfo{author}{P.~von Platen}, \bibinfo{author}{C.~Ma},
  \bibinfo{author}{Y.~Jernite}, \bibinfo{author}{J.~Plu},
  \bibinfo{author}{C.~Xu}, \bibinfo{author}{T.~L. Scao},
  \bibinfo{author}{S.~Gugger}, \bibinfo{author}{M.~Drame},
  \bibinfo{author}{Q.~Lhoest}, \bibinfo{author}{A.~M. Rush},
  \bibinfo{title}{Huggingface's transformers: State-of-the-art natural language
  processing}, \bibinfo{year}{2020}.
  \href{http://arxiv.org/abs/1910.03771}{{\tt arXiv:1910.03771}}.
\bibitem[{Kingma and Ba(2017)}]{kingma2017adam}
\bibinfo{author}{D.~P. Kingma}, \bibinfo{author}{J.~Ba}, \bibinfo{title}{Adam:
  A method for stochastic optimization}, \bibinfo{year}{2017}.
  \href{http://arxiv.org/abs/1412.6980}{{\tt arXiv:1412.6980}}.
\bibitem[{Zhang and Sabuncu(2018)}]{zhang2018generalized}
\bibinfo{author}{Z.~Zhang}, \bibinfo{author}{M.~R. Sabuncu},
  \bibinfo{title}{Generalized cross entropy loss for training deep neural
  networks with noisy labels}, \bibinfo{year}{2018}.
  \href{http://arxiv.org/abs/1805.07836}{{\tt arXiv:1805.07836}}.
\bibitem[{Agarap(2019)}]{agarap2019deep}
\bibinfo{author}{A.~F. Agarap}, \bibinfo{title}{Deep learning using rectified
  linear units (relu)}, \bibinfo{year}{2019}.
  \href{http://arxiv.org/abs/1803.08375}{{\tt arXiv:1803.08375}}.
\bibitem[{Merity et~al.(2017)Merity, Keskar, and
  Socher}]{merity2017regularizing}
\bibinfo{author}{S.~Merity}, \bibinfo{author}{N.~S. Keskar},
  \bibinfo{author}{R.~Socher}, \bibinfo{title}{Regularizing and optimizing lstm
  language models}, \bibinfo{year}{2017}.
  \href{http://arxiv.org/abs/1708.02182}{{\tt arXiv:1708.02182}}.
\bibitem[{Hande et~al.(2021)Hande, Puranik, Yasaswini, Priyadharshini,
  Thavareesan, Sampath, Shanmugavadivel, Thenmozhi, and
  Chakravarthi}]{hande2021offensive}
\bibinfo{author}{A.~Hande}, \bibinfo{author}{K.~Puranik},
  \bibinfo{author}{K.~Yasaswini}, \bibinfo{author}{R.~Priyadharshini},
  \bibinfo{author}{S.~Thavareesan}, \bibinfo{author}{A.~Sampath},
  \bibinfo{author}{K.~Shanmugavadivel}, \bibinfo{author}{D.~Thenmozhi},
  \bibinfo{author}{B.~R. Chakravarthi}, \bibinfo{title}{Offensive language
  identification in low-resourced code-mixed dravidian languages using
  pseudo-labeling}, \bibinfo{year}{2021}.
  \href{http://arxiv.org/abs/2108.12177}{{\tt arXiv:2108.12177}}.
\bibitem[{Sherstinsky(2020)}]{Sherstinsky_2020}
\bibinfo{author}{A.~Sherstinsky},
\newblock \bibinfo{title}{Fundamentals of recurrent neural network (rnn) and
  long short-term memory (lstm) network},
\newblock \bibinfo{journal}{Physica D: Nonlinear Phenomena}
  \bibinfo{volume}{404} (\bibinfo{year}{2020}) \bibinfo{pages}{132306}.
  \URLprefix \url{http://dx.doi.org/10.1016/j.physd.2019.132306}.
  \DOIprefix\doi{10.1016/j.physd.2019.132306}.
\bibitem[{Vaswani et~al.(2017)Vaswani, Shazeer, Parmar, Uszkoreit, Jones,
  Gomez, Kaiser, and Polosukhin}]{vaswani2017attention}
\bibinfo{author}{A.~Vaswani}, \bibinfo{author}{N.~Shazeer},
  \bibinfo{author}{N.~Parmar}, \bibinfo{author}{J.~Uszkoreit},
  \bibinfo{author}{L.~Jones}, \bibinfo{author}{A.~N. Gomez},
  \bibinfo{author}{{\L}.~Kaiser}, \bibinfo{author}{I.~Polosukhin},
\newblock \bibinfo{title}{Attention is all you need},
\newblock in: \bibinfo{booktitle}{Advances in neural information processing
  systems}, \bibinfo{year}{2017}, pp. \bibinfo{pages}{5998--6008}.
\bibitem[{Bahdanau et~al.(2016)Bahdanau, Cho, and Bengio}]{bahdanau2016neural}
\bibinfo{author}{D.~Bahdanau}, \bibinfo{author}{K.~Cho},
  \bibinfo{author}{Y.~Bengio}, \bibinfo{title}{Neural machine translation by
  jointly learning to align and translate}, \bibinfo{year}{2016}.
  \href{http://arxiv.org/abs/1409.0473}{{\tt arXiv:1409.0473}}.
\bibitem[{Thomas and Latha(2020)}]{thomas2020sentimental}
\bibinfo{author}{M.~Thomas}, \bibinfo{author}{C.~Latha},
\newblock \bibinfo{title}{Sentimental analysis of transliterated text in
  malayalam using recurrent neural networks},
\newblock \bibinfo{journal}{Journal of Ambient Intelligence and Humanized
  Computing}  (\bibinfo{year}{2020}) \bibinfo{pages}{1--8}.
\bibitem[{Ramesh et~al.(2021)Ramesh, Doddapaneni, Bheemaraj, Jobanputra, AK,
  Sharma, Sahoo, Diddee, J, Kakwani, Kumar, Pradeep, Deepak, Raghavan,
  Kunchukuttan, Kumar, and Khapra}]{ramesh2021samanantar}
\bibinfo{author}{G.~Ramesh}, \bibinfo{author}{S.~Doddapaneni},
  \bibinfo{author}{A.~Bheemaraj}, \bibinfo{author}{M.~Jobanputra},
  \bibinfo{author}{R.~AK}, \bibinfo{author}{A.~Sharma},
  \bibinfo{author}{S.~Sahoo}, \bibinfo{author}{H.~Diddee},
  \bibinfo{author}{M.~J}, \bibinfo{author}{D.~Kakwani},
  \bibinfo{author}{N.~Kumar}, \bibinfo{author}{A.~Pradeep},
  \bibinfo{author}{K.~Deepak}, \bibinfo{author}{V.~Raghavan},
  \bibinfo{author}{A.~Kunchukuttan}, \bibinfo{author}{P.~Kumar},
  \bibinfo{author}{M.~S. Khapra}, \bibinfo{title}{Samanantar: The largest
  publicly available parallel corpora collection for 11 indic languages},
  \bibinfo{year}{2021}. \href{http://arxiv.org/abs/2104.05596}{{\tt
  arXiv:2104.05596}}.
\bibitem[{Ott et~al.(2019)Ott, Edunov, Baevski, Fan, Gross, Ng, Grangier, and
  Auli}]{ott2019fairseq}
\bibinfo{author}{M.~Ott}, \bibinfo{author}{S.~Edunov},
  \bibinfo{author}{A.~Baevski}, \bibinfo{author}{A.~Fan},
  \bibinfo{author}{S.~Gross}, \bibinfo{author}{N.~Ng},
  \bibinfo{author}{D.~Grangier}, \bibinfo{author}{M.~Auli},
  \bibinfo{title}{fairseq: A fast, extensible toolkit for sequence modeling},
  \bibinfo{year}{2019}. \href{http://arxiv.org/abs/1904.01038}{{\tt
  arXiv:1904.01038}}.
\bibitem[{Puranik et~al.(2021)Puranik, Hande, Priyadharshini, Durairaj,
  Sampath, Thamburaj, and Chakravarthi}]{Puranik2021AttentiveFO}
\bibinfo{author}{K.~Puranik}, \bibinfo{author}{A.~Hande},
  \bibinfo{author}{R.~Priyadharshini}, \bibinfo{author}{T.~Durairaj},
  \bibinfo{author}{A.~Sampath}, \bibinfo{author}{K.~Thamburaj},
  \bibinfo{author}{B.~R. Chakravarthi},
\newblock \bibinfo{title}{Attentive fine-tuning of transformers for translation
  of low-resourced languages @loresmt 2021},
\newblock \bibinfo{year}{2021}.
\bibitem[{Papineni et~al.(2002)Papineni, Roukos, Ward, and Zhu}]{article23}
\bibinfo{author}{K.~Papineni}, \bibinfo{author}{S.~Roukos},
  \bibinfo{author}{T.~Ward}, \bibinfo{author}{W.~J. Zhu},
\newblock \bibinfo{title}{Bleu: a method for automatic evaluation of machine
  translation}  (\bibinfo{year}{2002}).
  \DOIprefix\doi{10.3115/1073083.1073135}.
\bibitem[{Kumar et~al.(2017)Kumar, Cotterell, Padró, and
  Oliver}]{inproceedings25}
\bibinfo{author}{A.~Kumar}, \bibinfo{author}{R.~Cotterell},
  \bibinfo{author}{L.~Padró}, \bibinfo{author}{A.~Oliver},
\newblock \bibinfo{title}{Morphological analysis of the dravidian language
  family},
\newblock \bibinfo{year}{2017}, pp. \bibinfo{pages}{217--222}.
  \DOIprefix\doi{10.18653/v1/E17-2035}.

\end{thebibliography}

\end{document}